\documentclass[letterpaper, 10 pt, conference]{ieeeconf}  
\IEEEoverridecommandlockouts                              
\pdfoutput=1                                              

\usepackage{algorithm}
\usepackage{algpseudocode}
\usepackage{amssymb,amsmath}
\usepackage[dvips]{color}
\usepackage{hyperref}
\usepackage{subfig}
\usepackage{array}
\usepackage{wasysym}
\usepackage{booktabs}
\usepackage[compress,nospace]{cite}
\usepackage{url}

\newif\ifpdf
\ifx\pdfoutput\undefined
   \pdffalse
\else
   \pdfoutput=1
   \pdftrue
\fi \ifpdf
   \usepackage{graphicx}
   \usepackage{epstopdf}
\else
   \usepackage{graphicx}
\fi

\title{\LARGE \bf
Control of an Aerial Manipulator using On-line Parameter Estimator for an Unknown Payload
}

\author{Hyeonbeom Lee$^{1}$, Suseong Kim$^{1}$ and H. Jin Kim$^{2}$
\thanks{${*}$This work was supported by the National Research Foundation of Korea (NRF) grant funded by the Korean government (MSIP) (2014R1A2A1A12067588).}
\thanks{${}^{1}$Hyeonbeom Lee and Suseong Kim are graduate students of Mechanical and Aerospace Engineering, Seoul National University, Seoul, Korea. {\tt\tiny koreaner33, suseongkim at snu.ac.kr}}%
\thanks{${}^{2}$H. Jin Kim is faculty in the Mechanical and Aerospace Engineering, Seoul National University, Seoul, Korea. {\tt\tiny hjinkim at snu.ac.kr}}%
}

\begin{document}

\maketitle
\thispagestyle{empty}
\pagestyle{empty}

\begin{abstract}
This paper presents an estimation and control
algorithm for an aerial manipulator using a hexacopter with
a 2-DOF robotic arm. The unknown parameters of a payload
are estimated by an on-line estimator based on parametrization
of the aerial manipulator dynamics. With the estimated
mass information and the augmented passivity-based controller,
the aerial manipulator can fly with the unknown object.
Simulation for an aerial manipulator is performed to
compare estimation performance between the proposed control
algorithm and conventional adaptive sliding mode controller.
Experimental results show a successful flight of a custom-made
aerial manipulator while the unknown parameters related to
an additional payload were estimated satisfactorily.
\end{abstract}


\section{Introduction} \label{sec:intro}
Mobile manipulation is a key to many applications of robotics such as construction sites, production lines and space environments. Among them, aerial manipulation is receiving attention for aerial transportation or inspection of hard-to-reach structures due to the superior mobility in three dimensional Euclidean space \cite{d2014guest}. 

Aerial manipulation can be divided into two categories based on the connection mechanism to a payload. The first approaches suspend a payload via a cable \cite{michael2011cooperative}. In \cite{michael2011cooperative}, they developed a control logic for equilibrium of the payload at a specific desired pose using three aerial manipulators. However, the possible pose of a payload is limited. The second approaches are to grasp and move the object by using a robotic arm \cite{lippiello2012exploiting,yang2014ams, fumagalli2012modeling, jimenez2013control}. In \cite{lippiello2012exploiting}, a dynamic model was derived by the Euler-Lagrangian formulation and cartesian impedance controller was designed for aerial manipulators. A tracking control law for the end effector was designed and tested in simulation based on decoupled Lagrangian dynamics \cite{yang2014ams}. In \cite{fumagalli2012modeling}, an aerial manipulator was developed for remote safety inspection of industrial plants. In \cite{jimenez2013control}, they custom-made an aerial manipulator with a 2-DOF robotic arm and designed integral backstepping controller. However, these papers do not consider the effect of unknown payload.

Handling of uncertain objects has been previously investigated for stationary manipulators or ground robots. \cite{slotine1987adaptive,liu2014passivity,dong2002trajectory,nestinger2012adaptive}. In \cite{slotine1987adaptive,liu2014passivity}, they presented an adaptive controller for a ground robotic manipulator to handle an unknown object. Control of a constrained ground manipulator with parameter uncertainty was shown in \cite{dong2002trajectory}. They designed an adaptive controller for trajectory and force tracking problems. The mass and inertia properties of a simple mobile robot are estimated by an adaptive estimator in \cite{nestinger2012adaptive}.

Research for aerial manipulators to handle an unknown object is still rare \cite{kim2013aerial,palunko2012trajectory,ruggiero2014impedance}. In \cite{kim2013aerial}, an adaptive sliding mode controller for an aerial manipulator was designed for coping with the parametric uncertainties. They demonstrated picking up and delivering an object with custom-made aerial manipulator. However, there was no estimation results and analysis of safe operation points for an unknown object. In \cite{palunko2012trajectory}, they proposed trajectory optimization and control for a quadrotor with the suspended payload. For handling a load, they used conventional adaptive control as same with \cite{slotine1987adaptive}, but this method is weak to noise or disturbances because the adaptation rule only based on control error. In \cite{ruggiero2014impedance}, they designed an estimator of external generalized forces  acting on aerial robots. This algorithm is applicable for aerial manipulator, but the performance for moving an unknown object has not been demonstrated by experiments.

There are researches on control of mechanical manipulators or aerial vehicles including estimation of unknown mass \cite{atkeson1986estimation,mellinger2011design}. In \cite{atkeson1986estimation}, they estimated the physical properties of a mechanical manipulator using least-squares method based on sensing of joint torque. However, this conventional estimation approach cannot be applied to small aerial manipulators, because the payload limitation does not allow to equip the heavy torque sensors.In \cite{mellinger2011design}, they developed a quadrotor with a gripper and used batch least-square method for estimating unknown mass using control input and acceleration data only. The estimation algorithm therein was designed for a quadrotor in hover or near-hover condition, but the dynamics will become more complicated for an aerial manipulator with a robotic arm. 

\begin{figure}
	\centering
	\includegraphics[width=0.7\columnwidth]{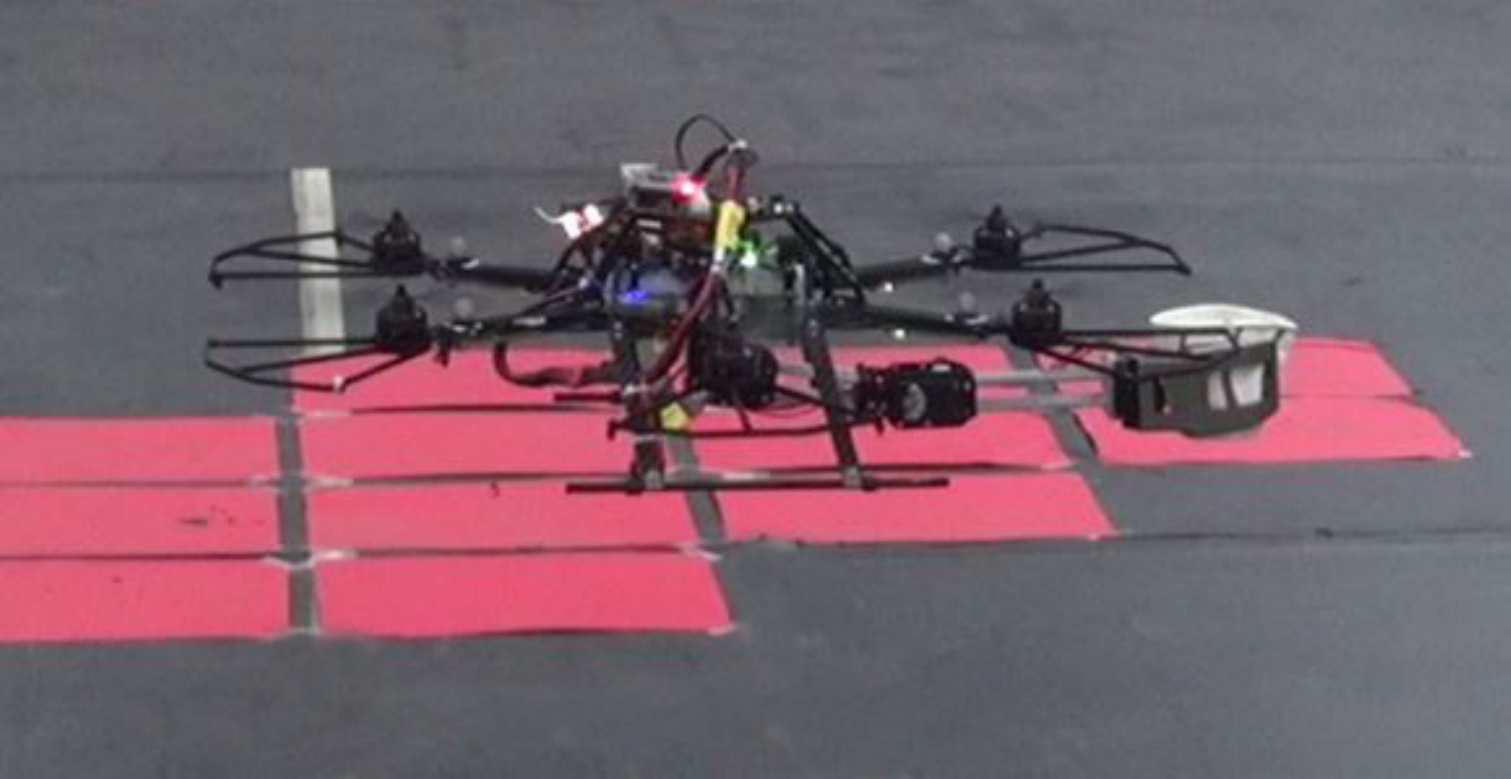}
	\caption{\textrm{\small Aerial manipulator using a hexacopter with a 2-DOF robotic arm.}}
	\label{fig:hexa}
\end{figure}

The contributions of this paper can be summarized as follows: First, we design an on-line parameter estimator for the aerial manipulator. The unknown properties of a payload are estimated by the parameter estimator based on parametrization of the combined system, which considers a hexcopter and a robotic arm as a unified system. Unlike \cite{mellinger2011design}, the estimator does not depend on the flight condition. Second, an augmented passivity-based controller is designed to stabilize the aerial manipulator with the estimated parameters. Third, we apply this control algorithm to our custom-made aerial manipulator as shown in Fig. \ref{fig:hexa} and show feasibility of the proposed control algorithm for the aerial manipulator in the simulation and experiment.

This paper is structured as follows: in section \ref{sec:quad_dyn}, we describe the dynamics of an aerial manipulator. The parameter estimator and controller are designed in section \ref{sec:design}. Section \ref{sec:exp} presents results from experiments in which the aerial manipulator estimates the unknown parameters of a payload on the flight. Section \ref{sec:con} contains concluding remarks.

\section{Dynamics of Aerial Manipulator} \label{sec:quad_dyn}
This section presents the dynamics of aerial manipulator. The more detailed kinematic relations and equations of motion can be found in our previous research \cite{kim2013aerial}.

\begin{figure}
	\centering
	\includegraphics[width=0.7\columnwidth]{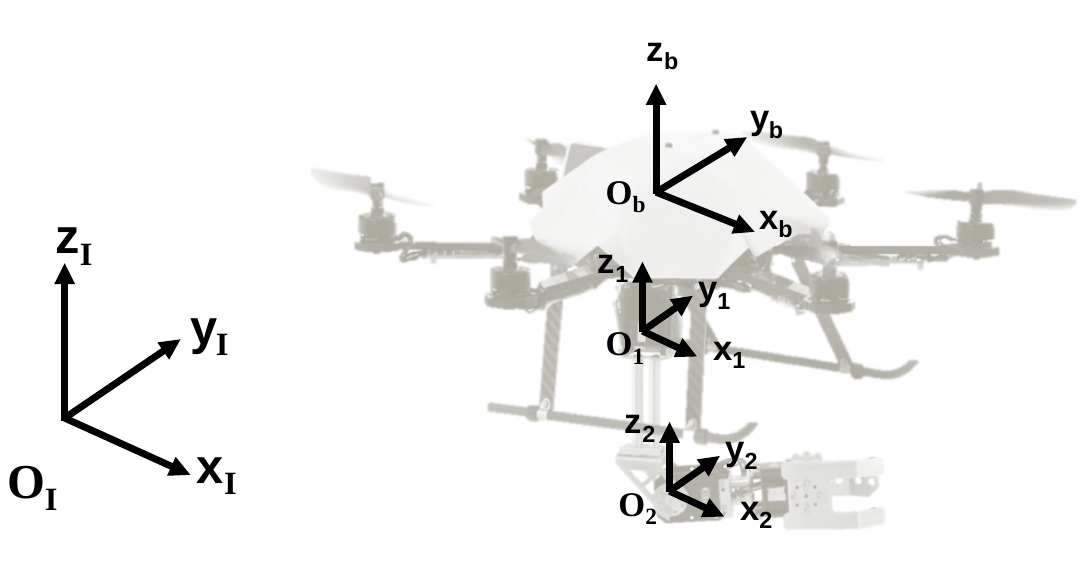}
	\caption{\textrm{\small Configuration of the coordinates for the combined system consisting of a hexacopter and a 2-DOF arm.}}
	\label{fig:config}
\end{figure}

\subsection{Dynamics for the Combined System}
Fig. \ref{fig:config} shows coordinated frames for the dynamic model of an aerial manipulator. $O_I, O_b, O_i$ represent the inertial frame, the bodyframe of the hexacopter and the bodyframe of link $i$, respectively. The subscript $i=1,2$ denotes the link number. Using the position of center of mass of the hexacopter in the inertial frame $p=[x,y,z]^T$, Euler angles of the hexacopter $\Phi=[\phi,\theta,\psi]^T$ and joint angles of the manipulator $\eta=[\eta_1, \eta_2]^T$, the dynamic model can be described based on the following system state,
\begin{align}
q = \left[ {\begin{array}{*{20}{c}}
p^T&\Phi^T &\eta^T
\end{array}} \right]^T.
\end{align}

The equations of motion of the combined system with the state $q$ can be derived as
\begin{align}
M(q)\ddot q + C(q,\dot q)\dot q + G(q) = \tau  \label{eqn:mcg_model},
\end{align}
where $\tau\in\mathbb{R}^{8\times1}$ is the control input, $M(q)\in\mathbb{R}^{8\times8}$ is the inertia matrix, $C(q,\dot q)\in\mathbb{R}^{8\times8}$ is the Coriolis matrix, and $G(q)\in\mathbb{R}^{8\times1}$ is the gravity term. Here, two elements of $\tau$, i.e. $\tau(1)$ and $\tau(2)$, are used to generate the desired roll $\phi_d$ and pitch angle $\theta_d$. They are computed by the following rule:
\begin{align}
\left[ {\begin{array}{*{20}{c}}
{{\theta_d}}\\
{{\phi _d}}
\end{array}} \right] = \frac{1}{\tau(3)}\left[ {\begin{array}{*{20}{c}}
{\cos (\psi )}&{\sin (\psi )}\\
{\sin (\psi )}&{ - \cos (\psi )}
\end{array}} \right]\left[ {\begin{array}{*{20}{c}}
{\tau (1)}\\
{\tau (2)}
\end{array}} \right]. \label{eqn:phid}
\end{align}
This relation is derived based on the small roll and pitch angle assumption.

\subsection{System Parametrization}

\begin{figure}
	\centering
	\includegraphics[width=0.7\columnwidth]{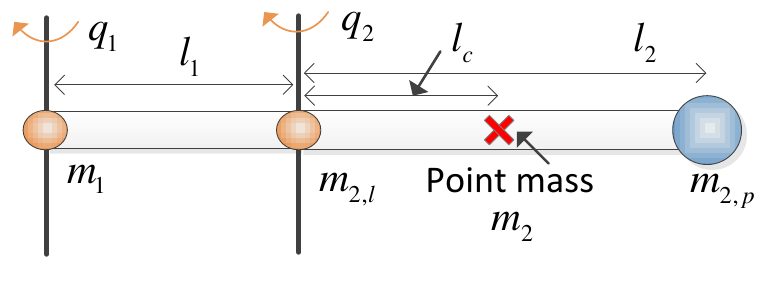}
	\caption{\textrm{\small An additional payload is attached to the end-effector.  ($m_1,m_{2,l},m_{2,p}$ : mass of link 1, link 2, additional payload, respectively)}}
	\label{fig:change_inertia}
\end{figure}
When the end effector grasps a single object, the physical properties of link 2 change. Total mass of second link, $m_2$ is changed by additional payload ${m_{2,p}}$ and initial mass of second link $m_{2,l}$, i.e. $m_2=m_{2,l}+m_{2,p}$ as shown in Fig. \ref{fig:change_inertia}. Also, if the additional payload is treated as a point mass, then the change of moment of inertia is easily computed by the parallel axis theorem \cite{hibbeler2001engineering}:
\begin{align}
I_{y,2}^\ast=I_{y,2}+m_{2,p}l_{2}^2,
\end{align}
where $I_{y,2}^\ast$ is the changed moment of inertia with respect to rotational axis of second link and $l_2$ is length of second link.

Since $m_{2,l}$ and other physical information about hexacopter and link 1 are already known values, therefore, the equation of motion is parameterized with respect to the unknown mass $m_2$ and the length of center of mass at link 2, $l_c$. Here, $m_2$ is the total mass of link 2 including the unknown payload. In this case, we can separate $M$, $C$ and $G$ into the matrices with known parameters and unknown parameters:
\begin{align}
M&=M_1+{m_2}M_2+{m_3}M_3+{m_4}M_4 \nonumber \\
C&=C_1+{m_2}C_2+{m_3}C_3+{m_4}C_4  \nonumber\\
G&=G_1+{m_2}G_2+{m_3}G_3 \label{eqn:mcg_param}.
\end{align}
Here $M_1$, $C_1$ and $G_1$ are the matrices with known physical parameters and the others are the matrices with unknown parameters, such as unknown mass $m_2$ and $m_3=m_2l_c$, $m_4={m_2}l_c^2$. Using these matrices \eqref{eqn:mcg_param}, the dynamic equations \eqref{eqn:mcg_model} can be written as the following parameterized form:
\begin{align}
M_1\ddot q +C_1\dot q+G_1+{m_2}({M_2}\ddot q+ {C_2}\dot q + {G_2}) \nonumber \\+ {m_3}({M_3}\ddot q + {C_3}\dot q + {G_3}) + {m_4}({M_4}\ddot q + C_4\dot q)= \tau \label{eqn:para_pre}
\end{align}

For simplicity, we can rewrite \eqref{eqn:para_pre} as
\begin{align}
{m_2}({M_2}\ddot q+ {C_2}\dot q + {G_2})+ {m_3}({M_3}\ddot q + {C_3}\dot q + {G_3})\nonumber \\ + {m_4}({M_4}\ddot q + C_4 \dot q)= U(t) \label{eqn:paramtri},
\end{align}
by introducing the the forcing term including control input $\tau$ in \eqref{eqn:paramtri}:
\begin{align}
U(t) := \tau - {M_1}\ddot q(t)- {C_1}\dot q(t) - G_1.
\end{align} 
Here, forcing term $U(t)$ will be exploited in parameter estimator, which will be explained more detail in the next section.
\section{Estimator and Controller Design} \label{sec:design}
\begin{figure}
	\centering
	\includegraphics[width=0.7\columnwidth]{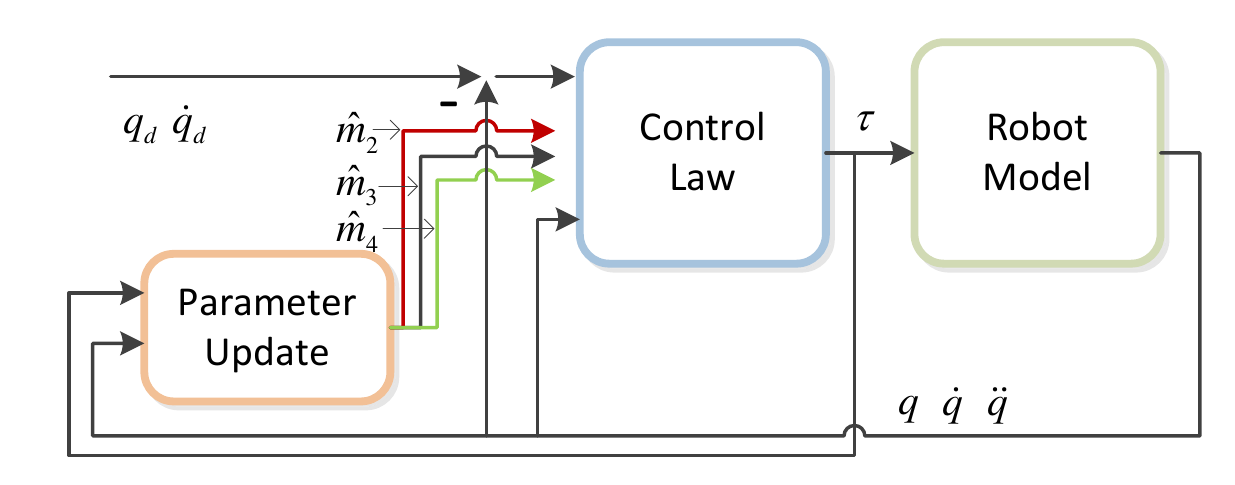}
	\caption{\textrm{\small Proposed control structure for the quadrotor with robotic arm.}}
	\label{fig:structure}
\end{figure}

In this section, we design on-line parameter estimator and the augmented passivity-based controller for the hexacopter with the robotic arm. The total control structure is shown as Fig. \ref{fig:structure}.

\subsection{On-line Parameter Estimator for an Unknown Payload}
In order to estimate the physical properties of an unknown payload, a parameter estimation law for an aerial manipulator is developed based on the parametrization of dynamic equation \eqref{eqn:paramtri}. For a parameter estimator, we assumed that $m_2$, $m_3$ and $m_4$ are constant parameters to be identified on-line. We also assumed that $q$, $\dot q$ and $\ddot q$ are available with the forcing term $U(t)$ which makes $q$ bounded. Note that the second assumption can be relaxed by the control law. Based on these assumptions, the state estimator equation appears as
\begin{align}
&C^* \dot {\hat q}(t)  + K^* \hat q(t) + [{\hat m_2}(t){M_2} + \hat m_3(t){M_3}+
\hat m_4(t){M_4}]\ddot q(t) \nonumber \\&+ [{\hat m_2}(t){C_2} + \hat m_3(t){C_3}+
\hat m_4(t){C_4}]\dot q(t) + \hat m_2G_2+\hat m_3G_3 \nonumber \\&= U(t) + {C^*}\dot q(t) + K^* q(t), \label{eqn:estimator}
\end{align}
where $C^*$ and $K^*$ are user-defined positive definite gain matrices. Here, $\hat m_2$, $\hat m_3$ and $\hat m_4$. are the estimated parameters. Note that ${\dot {\hat q}}(0) \ne \dot q(0)$ should be satisfied for estimation.

If we define the state error as
\begin{align}
e(t) = {\hat q} (t) - q(t) \label{eqn:error},
\end{align}
where ${\hat q}(t)$ is the estimated state. The parameter update rules can be given as
\begin{align}
\dot {\hat m}_2(t)&= {\gamma _1}{ e^T}(t)({M_2}\ddot q(t)+{C_2}\dot q(t)+G_2) \nonumber \\
\dot {\hat m}_3(t)&= {\gamma _2}{ e^T}(t)({M_3}\ddot q(t)+{C_3}\dot q(t)+G_3) \nonumber \\
\dot {\hat m}_4(t) &= {\gamma _3}{e^T}(t)({M_4}\ddot q(t)+{C_4}\dot q(t)) \label{eqn:update},
\end{align}
where, the positive numbers $\gamma_1, \gamma_2, \gamma_3$ are the learning rate of parameter estimator.

Then, using \eqref{eqn:estimator} and \eqref{eqn:error}, estimation error dynamics can be written as:
\begin{align}
{C^*}\dot e(t) + {K^*}e(t) + ({\tilde m_2}{M_2} + {\tilde m_3}{M_3} + {\tilde m_4}{M_4})\ddot q \nonumber \\({\tilde m_2}{C_2} + {\tilde m_3}{C_3} + {\tilde m_4}{C_4})\dot q + \tilde m_2G_2+\tilde m_3G_3 = 0. \label{eqn:esterrdyn}
\end{align}

\noindent \textbf{Proposition:} The error dynamics \eqref{eqn:esterrdyn} is asymptotically stable.

\begin{proof}
In order to prove the convergence of the estimation error dynamics \eqref{eqn:estimator}, we define the following Lyapunov candidate function:
\begin{align}
V_1 = \frac{1}{2}e(t)^T{C^*}e(t) + \frac{1}{{2{\gamma _1}}}\tilde m_2^2(t) +\frac{1}{{2{\gamma _2}}}\tilde m_3^2(t) + \frac{1}{{2{\gamma _3}}}\tilde m_4^2(t).
\end{align} The time derivative of $V_1$ is given as:
\begin{align}
&\dot V_1 =  e^T(t){C^*}\dot e(t) + \frac{1}{{{\gamma _1}}}{\tilde m_2}{\dot {\hat m}}_2 + \frac{1}{{{\gamma _2}}}{\tilde m_3}{\dot {\hat m}}_3 + \frac{1}{{{\gamma _3}}}{\tilde m_4}{\dot {\hat m}}_4 \nonumber \\
=&  - {e^T}(t){K^*}e(t) - {e^T}(t)[{\tilde m_2}(t){M_2} + {\tilde m_3}(t){M_3} \nonumber \\&+ {\tilde m_4}(t){M_4}]\ddot q(t) - {e^T}(t)[{\tilde m_2}(t){C_2} + {\tilde m_3}(t){C_3} \nonumber \\&+ {\tilde m_4}(t){C_4}]\dot q(t) - {e^T}(t)[{\tilde m_2}(t){G_2} + {\tilde m_3}(t){G_3}]{\rm{ }}  \nonumber \\&+ \frac{1}{{{\gamma _1}}}{\tilde m}_2(t){\dot {\hat m}}_2(t) + \frac{1}{{{\gamma _2}}}{\tilde m}_3(t){\dot {\hat m}}_3(t) + \frac{1}{{{\gamma _3}}}{\tilde m}_4(t){\dot {\hat m}}_4(t)\nonumber \\
=& - {e^T}(t){K^*}e(t) + {\tilde m_2}(t)[\frac{{{{\dot {\hat m}}_2}(t)}}{{{\gamma _1}}} - {e^T}(t)({M_2}\ddot q(t) +{C_2}\dot q(t)\nonumber \\&+ {G_2})] + {\tilde m_3}(t)[\frac{{{{\dot {\hat m}}_3}(t)}}{{{\gamma _2}}} -{e^T}(t)({M_3}\ddot q(t)+{C_2}\dot q(t) + {G_3})] \nonumber \\&+ {\tilde m_4}(t)[\frac{{{{\dot {\hat m}}_4}(t)}}{{{\gamma _3}}} - {e^T}(t)({M_4}\ddot q(t)+{C_4}\dot q(t))] \nonumber.
\end{align}
Using the fact that ${\dot {\hat {m}}}_*={\dot {\tilde {m}}}_*$, the parameter estimator rule \eqref{eqn:update} results in:
\begin{align}
\dot V_1 = - {{ e(t)}^T}{K^*}e(t) \leq -\lambda_{min}(K^*)\parallel e(t)\parallel^2 \leq 0 \label{eqn:dv1},
\end{align}
where $\lambda_{min}(K^*)$ denotes the smallest eigenvalue of the matrix $K^*$. This proves the bounded of $e$, ${\tilde m}_2$, ${\tilde m}_3$, ${\tilde m}_4$. If $\ddot q$ is bounded, then $\dot e$ is bounded by \eqref{eqn:esterrdyn}. Then, $\ddot V_1=- 2{{ e}^T}(t){K^*}{\dot {e}}(t)$ is also bounded, which guarantees that state estimation error, $e(t)$ goes to 0 as time goes infinity by application of Barbalat's lemma \cite{khalil2002nonlinear}. 
\end{proof}

{The proof of convergence when persistence of excitation is assumed, i.e., $({\tilde m}_2,{\tilde m}_3,{\tilde m}_4)\to0$, follows from standard results in \cite{atkeson1986estimation}. The detailed proof is omitted due to length limitation.}

\subsection{Augmented Passivity-based Controller Design}
In this subsection, we design the controller for the aerial manipulator. Applying the estimated parameters $\hat m_2$, $\hat m_3$ and $\hat m_4$ to the controller, a augmented passivity-based control law is designed with following state error as
\begin{align}
{e_c}&= q - {q_d}.
\end{align}
{Here, we assumed that the desired trajectory is bounded as follows:
\begin{align}
\left|q_d\right|^2+\left|\dot{q}_d\right|^2+\left|\ddot{q}_d\right|^2\le\rho,
\end{align}
where $\rho$ is a positive constant.}

The augmented passivity-based control for robotic manipulators using the estimated parameters can be written as the following equation:
\begin{align}
\tau  = & {\hat M}( q){\ddot q_d}+{\hat C}( q){\dot q_d} +{{\hat G}}(q) - {k}({\dot e_c}+\Lambda{e_c}) \label{eqn:tau2}
\end{align}
where $k$ and $\Lambda$ are diagonal gain matrices.  ${{\hat M}}$, ${{\hat C}}$ and ${{\hat G}}$ represent the estimation of each matrix by parameter update rule \eqref{eqn:update}, i.e. ${\hat M}={M_1}( q) + {\hat m_2}{M_2}( q) + {\hat m_3}{M_3}(\hat q) + {\hat m_4}{M_4}(q)$, ${\hat C}={C_1}( q) + {\hat m_2}{C_2}( q) + {\hat m_3}{C_3}(\hat q) + {\hat m_4}{C_4}(q)$ and ${\hat G}={{\hat m}_2}G_2( q)+{{\hat m}_3}G_3(q)+G_1(q)$.

Before proving the stability, we define a regressor matrix, $Y\in \mathbb{R}^{8\times4}$ for simplicity and it appears as:
\begin{align}
\hat M {\ddot q}_d+\hat C {\dot q}_d+\hat G=Y(q,\dot q,{\dot q}_d,{\ddot q}_d)\hat \xi,
\end{align}
where $\xi=[m_2, m_3, m_4,1]^T$, $\hat \xi$ is the estimation of $\xi$ and
\begin{align}
\tiny
Y= [Y_1^T,Y_2^T,Y_3^T,Y_4^T] \label{eqn:reg},
\end{align}
where $Y_1={M_2}{{\ddot q}_d}+{C_2}{\dot q}_d  + {G_2}$, $Y_2={M_3}{{\ddot q}_d}+{C_3}{\dot q}_d  + {G_3}$, $Y_3={M_4}{{\ddot q}_d}+{C_4}{\dot q}_d$ and $Y_4={M_1}{{\ddot q}_d}+{C_1}{\dot q}_d  + {G_1}$.

The closed-loop dynamics can be derived by substituting the proposed control law \eqref{eqn:tau2} into \eqref{eqn:mcg_model} and using \eqref{eqn:reg}:
\begin{align}
{M}\ddot e_c + (C+{k})\dot e_c + {k\Lambda}e_c = Y(\hat \xi-\xi). \label{eqn:err_con}
\end{align}
With these equations, stability analysis can be performed as the following, which shows the boundedness of error $e_c$ and ${\dot e}_c$.

\begin{proof}
We define the following Lyapunov candidate function:
\begin{align}
{V_2} = \frac{1}{2}{\dot e_c^T}{M}\dot e_c + \frac{1}{2}{e_c^T}{k\Lambda}e_c >0.
\end{align}

The time derivative of $V_2$ can be expressed as:
\begin{align}
{\dot V_2}=& {\dot e_c^T}{M}\ddot e_c+\frac{1}{2}{\dot e_c^T}{\dot M}\dot e_c +{e_c^T}{k\Lambda}\dot e_c     \label{eqn:v2dot} \\
    =& \frac{1}{2}{\dot e_c^T}({\dot M}-2C)\dot e_c - {\dot e_c^T}{k}\dot e_c - {\dot e_c^T}{k\Lambda}e_c \nonumber \\&+ {\dot e}_c^TY(q,{\ddot q}_d)(\hat \xi-\xi)+{e_c^T}{k\Lambda}\dot e_c \nonumber \\
    =& - {\dot e_c^T}{k}\dot e_c+{\dot e}_c^TY(q,\dot q,{\dot q}_d,{\ddot q}_d)(\hat \xi-\xi),\nonumber
\end{align}
In the derivation, skew symmetricity of $({\dot M} - 2C)$ is used \cite{murray1994mathematical}. Although $\left\| {\hat \xi  - \xi } \right\| \neq 0$, the term $Y(q,\dot q,{\dot q}_d,{\ddot q}_d)(\hat \xi-\xi)$ is bounded by the parameter estimator. In this case, if we choose a gain $k$ sufficiently large, we can show that ${V_2}$ is ultimately bounded. However, if we have the parameter convergence under the proposed estimator in \eqref{eqn:estimator}, i.e., $\left\| {\hat \xi  - \xi } \right\| = 0$, we can show that ${\dot V_2}\leq0$. In this situation, as same with \eqref{eqn:dv1}, we can show asymptotic stability of the proposed controller by application of Barbalat’s lemma \cite{khalil2002nonlinear}. 
\end{proof}

\section{Simulation Results} \label{sec:simul}
Precise estimation of parameters is important for handling an unknown payload. Therefore, in this section, simulation for an aerial manipulator is performed to compare estimation performance between the proposed control algorithm and conventional adaptive sliding mode controller.

\subsection{Mass Estimation using Adaptive Sliding Mode Controller}
An adaptive sliding mode controller for an aerial manipulator is shown in \cite{kim2013aerial}.
For performance comparison with the proposed method, we also implemented a simple estimation algorithm described below, which is modified based on \cite{mellinger2011design} for an additional payload.

To design an adaptive sliding mode controller, we can define the sliding surface $s$ and virtual reference trajectory $q_r$ as
\begin{align}
s &= \dot{q}-\dot{q}_r \\
{\dot q}_r &= {\dot q_d} - \Lambda ( q - {{q}_d}), \nonumber
\end{align}
where $\Lambda$ is a diagonal gain matrix.

The adaptive sliding mode controller appears as
\begin{align}
\tau  = \hat {M}{\ddot q_r}+\hat {C}{\dot q_r} + \hat {G} + \hat \Delta  - K_1s - K_2{\mathop{\rm sgn}} (s), \label{eqn:asc1}
\end{align}
where $K_1$ and $K_2$ are the diagonal gain matrices and $\dot {\hat \Delta } =  - [(\dot q - {\dot q_d}) + \Lambda (q - {q_d})]$ is the adaptation term for cancelling out the modelling error and disturbances. Also, $\hat M$, $\hat C$ and $\hat G$ are the function of the states, i.e., $q$ and the estimated unknown mass ${\hat m}_2$.

We can extract the unknown mass information of a payload using the adaptive sliding mode controller by the following equation:
\begin{align}
\hat m &= \frac{{\tau(3)}}{g + {\hat M}_{3}(q,{\hat m}_2^-)\ddot q + {\hat C}_{3}(q,\dot q,{\hat m}_2^-)\dot q}\label{eqn:asc2} \\
{{\hat m}_2} &= \hat m - ({m_b} + {m_1}) \nonumber ,
\end{align}
where the subscript $*_{3}$ is the third row of the matrix $*$, ${\hat m}_2^-$ are previously estimated mass and  $m$ is total mass of the arial manipulator containing the unknown mass. Here $m_b$ and $m_1$, which are already known, are the mass of a hexacopter or link 1, respectively.

\subsection{Hexacopter with Robotic Arm Model}
The parameters of a hexacopter with two-DOF robotic arm under consideration are given in Table. \ref{tab:para}. Properties of the Firefly hexacopter are from \cite{achtelik2012design}. Here $l$ is the length of link $i (i=1,2)$. The mass of a payload is set to be $0.4 kg$, i.e., $m_2=0.5$ kg.

\begin{table}[h]
\caption{Firefly hexacopter with 2-DOF robotic arm model}
\label{tab:para}
\begin{center}
\begin{tabular}{|c||c||c||c|}
 \hline
 Parameter & Value & Parameter & Value\\
 \hline
 $m_b$ & 1.0 kg & $Ixx$ & 0.013 kgm$^2$\\
 \hline
  $Iyy$ & 0.013 kgm$^2$ &  $Izz$ & 0.021 kgm$^2$\\
 \hline
  $m_1, m_2^*$& 0.1kg & $l_1, l_2$& 0.2m \\
 \hline
\end{tabular}
\end{center}
\text{\,\,\,\,\,\,\,\,\,\,\,\,* : initial value. }
\end{table}

\subsection{Simulation Results}
Parameter estimation was simulated in time interval $[0,10]$ with $\gamma_1=0.2$, $\gamma_2=0.1$, $\gamma_1=0.1$, $C*=10 \times I_{8\times8}$, $K*=20 \times I_{8\times8}$. Gains for the proposed control law are set as $k=diag[4.5, 4.5, 7.5, 8.0, 8.0, 8.0, 1.4,  1.4]$, $\Lambda=diag[1.0, 1.0, 5.0, 1.0, 1.0, 1.0, 0.2, 0.2]$.

For the estimation of unknown parameters, we set the initial estimated state as ${\dot {\hat q}}(0)=0.1 \times I_{8\times8}$. We also set initial value of unknown parameters as ${\hat m}_2(0)=0.1$, $\hat m_3(0)=1.0\times10^{-3}$ and $\hat m_4(0)=1.0\times10^{-5}$. Additional payload is set to be $0.4$kg, i.e. ${m}_2=0.5$ and ${l}_c=0.16$.

The desired trajectory that the aerial manipulator should follow is set to be as
\begin{align}
{q_d} = \Big[&{\frac{1}{2}\cos (\frac{\pi }{5}t)},{ - \frac{1}{2}\cos (\frac{\pi }{5}t)},{0.7},{{\phi _d}},{{\theta _d}},0,\nonumber\\&{ - \frac{\pi }{2} + \frac{\pi }{4}\sin (\frac{\pi }{5}t)},{\frac{\pi }{8}\sin (\frac{\pi }{5}t)}\Big]^T,
\end{align}
where $\phi_d$ and $\theta_d$ are specified by \eqref{eqn:phid}.

\begin{figure}
	\centering
     {\includegraphics[width=0.55\columnwidth]{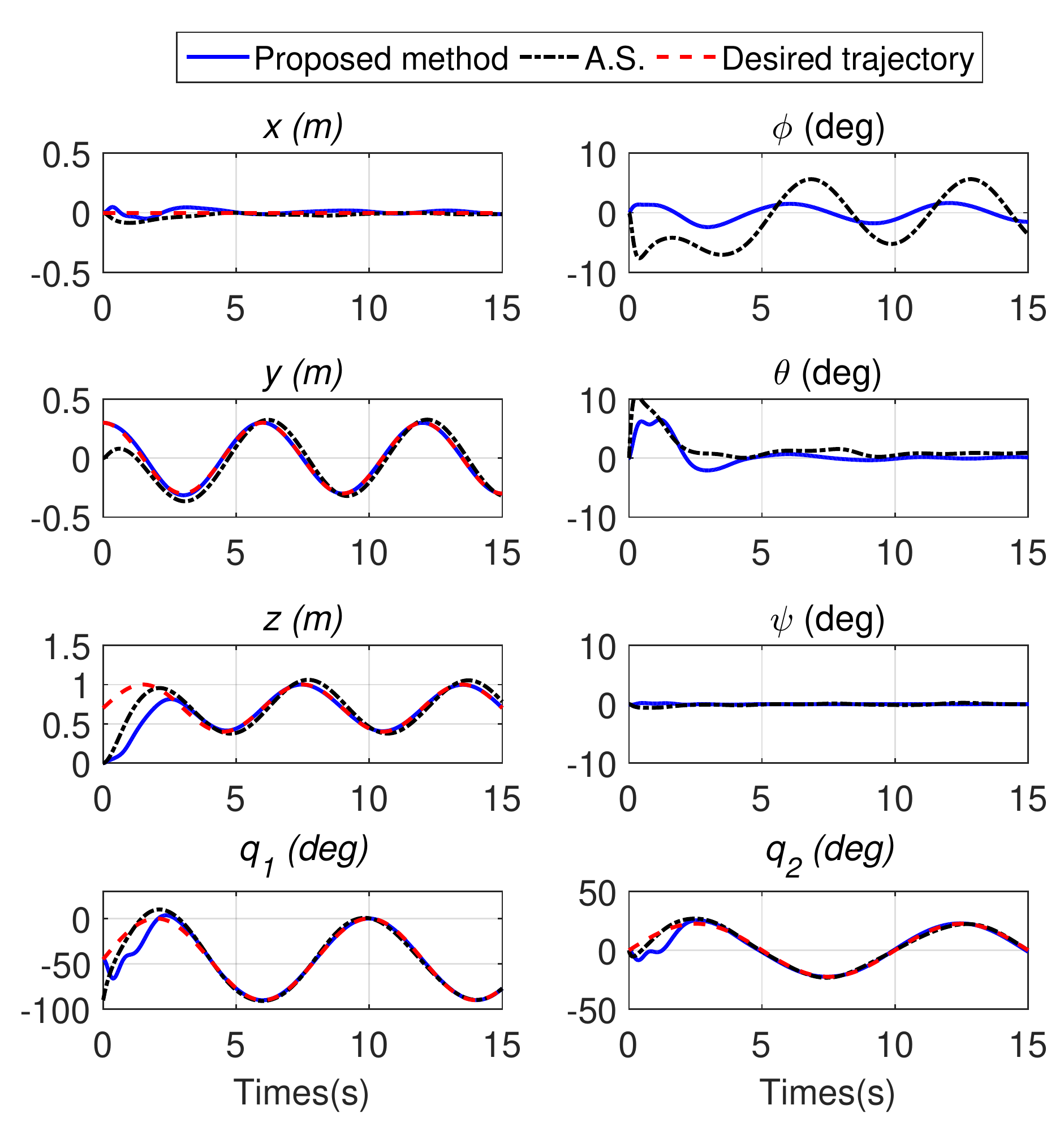}}
	\caption{\textrm{\small Performance of the trajectory tracking.}}
	\label{fig:posi_track}
\end{figure}

\begin{figure}
	\centering
    \subfloat[State estimation performance ($q$ vs. ${\hat q}$).]
	   {\includegraphics[width=0.5\columnwidth]{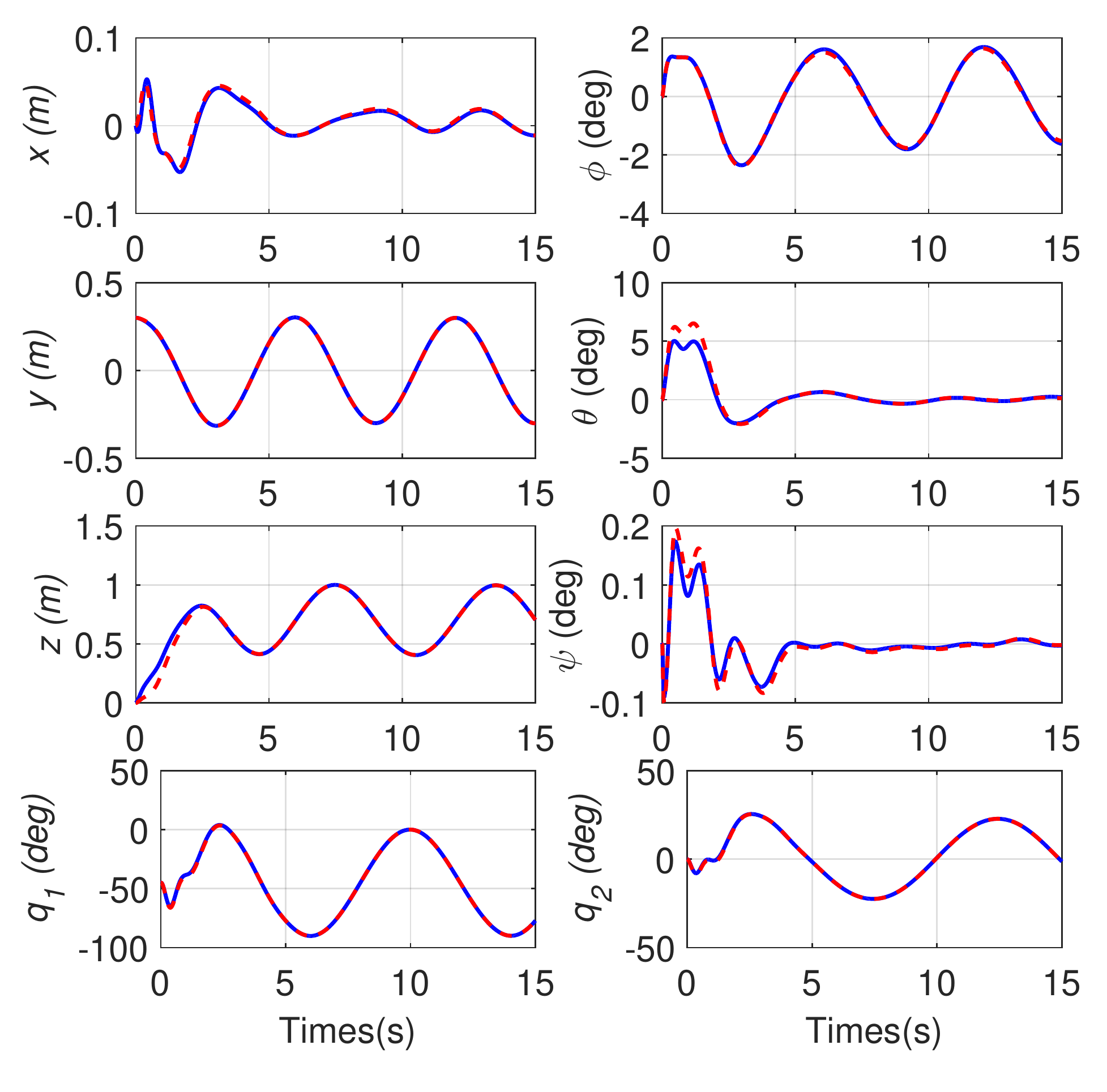}}
 	\subfloat[Parameter Estimation. \label{fig:est2}]
      {\includegraphics[width=0.5\columnwidth]{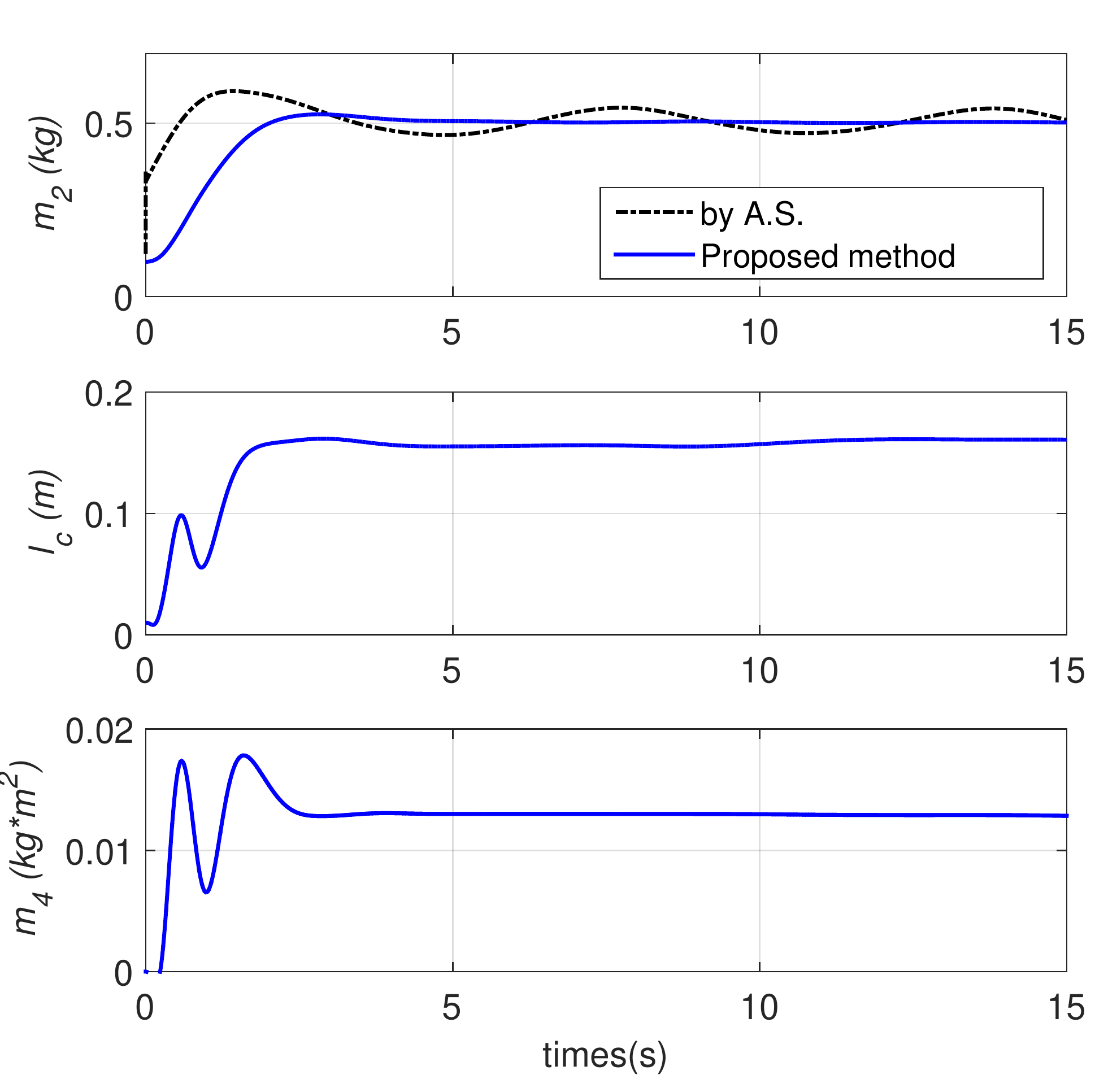}}
	\caption{\textrm{\small Estimation results during the flight shown in Fig. \ref{fig:posi_track}.}}
	\label{fig:est_per}
\end{figure}

Fig. \ref{fig:posi_track} shows the position tracking performance. The red dashed line is the desired trajectory and the blue solid line is the trajectory of the aerial manipulator using the proposed control algorithm. Fig. \ref{fig:est_per} shows the estimator performance and results of estimated parameters. The blue solid line is the state of the aerial manipulator and the red dashed line is the estimated state by the parameter estimator.

With the proposed control algorithm, the aerial manipulator satisfactorily tracks the desired trajectory even when there exists an unknown payload. The unknown parameters $m_2$, $m_3$ and $m_4$ are estimated precisely as shown in Fig. \ref{fig:est2}. However, when using the adaptive sliding mode controller \eqref{eqn:asc1} and \eqref{eqn:asc2}, the performance of estimation became worse than the proposed control law as shown in Fig. \ref{fig:est2}. This is mainly because the adaptive sliding mode control algorithm can estimate the unknown mass only while the parameter estimator estimate the whole unknown parameters, $m_2, m_3, m_4$.

\section{Experimental Results} \label{sec:exp}
In this section, we describe experimental results with a custom-made aerial manipulator, which is composed of a hexacopter and a 2-DOF robotic arm.

For facilitating easier implementation, we use the following simplicity assumptions:
\begin{itemize}
     \item Roll and pitch angles are small, i.e., $|\phi|\ll1$ and $|\theta|\ll1$.
     \item At link 1, center of mass is located at joint angle, i.e. $l_{c1}\approx0$.
\end{itemize}
Using these two assumption, the on-line parameter equation \eqref{eqn:estimator} and the proposed controller \eqref{eqn:tau2} can be implemented much simpler.

\subsection{Experimental Setup}
The hexacopter used in this paper is a Ascending Technlogies Firefly hexacopter \cite{asctec}. The robotic arm is customized with Dynamixel servomotors. The total length of arm is 0.305 meter: $l_1=0.135$ meter and $l_2=0.17$ meter. The total weight of robotic arm is about 0.220 kg before picking up the payload.

\begin{figure}
	\centering
	\includegraphics[width=0.7\columnwidth]{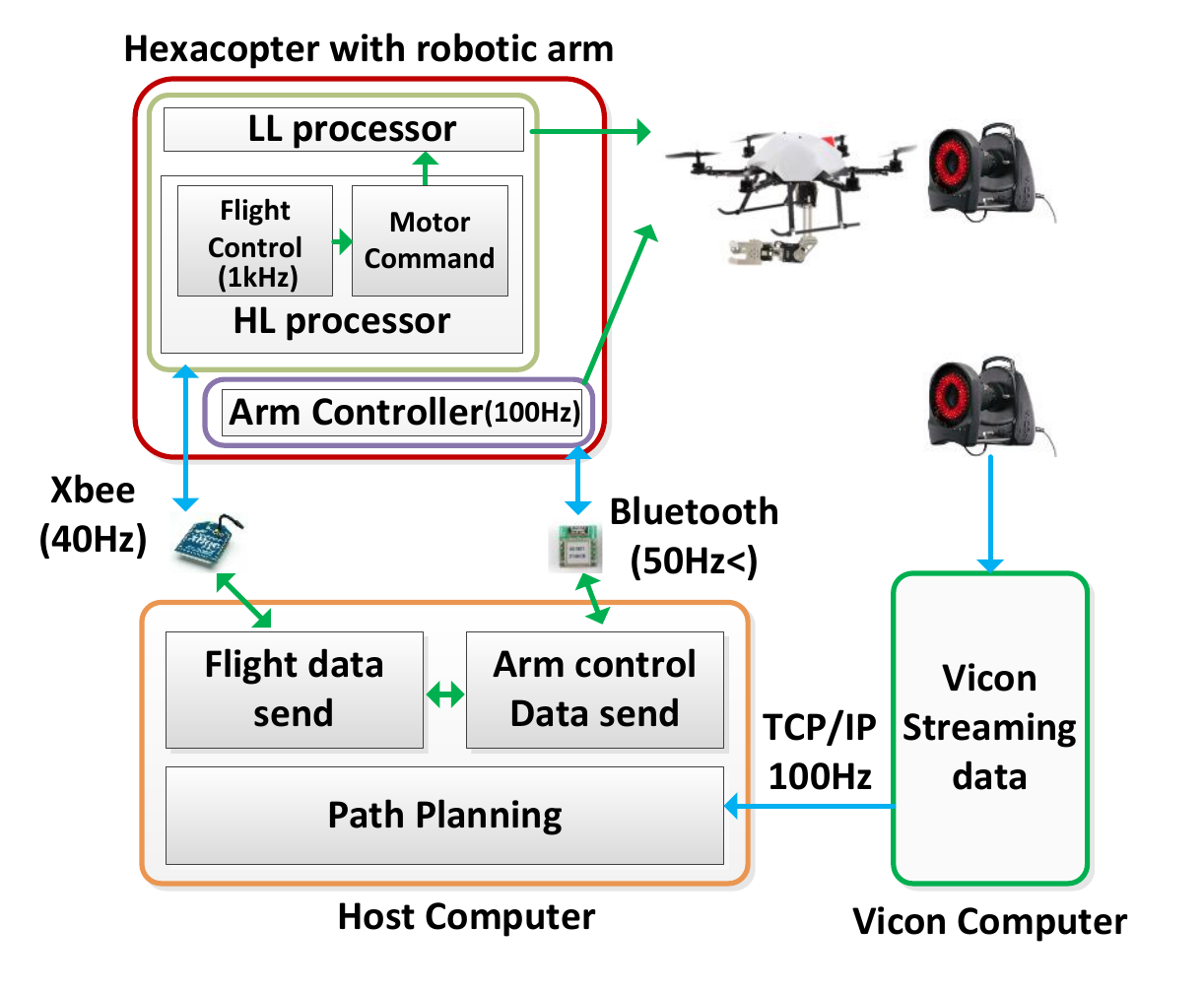}
	\caption{\textrm{\small Experimental setup.}}
	\label{fig:expset}
\end{figure}

The experimental setup is shown in Fig. \ref{fig:expset}. Vicon, an indoor GPS system, gives the position informations with 100 Hz to the base computer. The desired and current states of a hexacopter and the joint angles are transmitted to the hexaxcopter with Xbee at 40 Hz. The parameter estimator and augmented controller runs at 1 kHz in the onboard processor of Firefly hexacopter. Arm control inputs are sent by Bluetooh at 50 Hz.

\subsection{Experimental Results}

\begin{figure}
	\centering
    \subfloat[Tracking performances\label{fig:flight1}]
	   {\includegraphics[width=0.5\columnwidth]{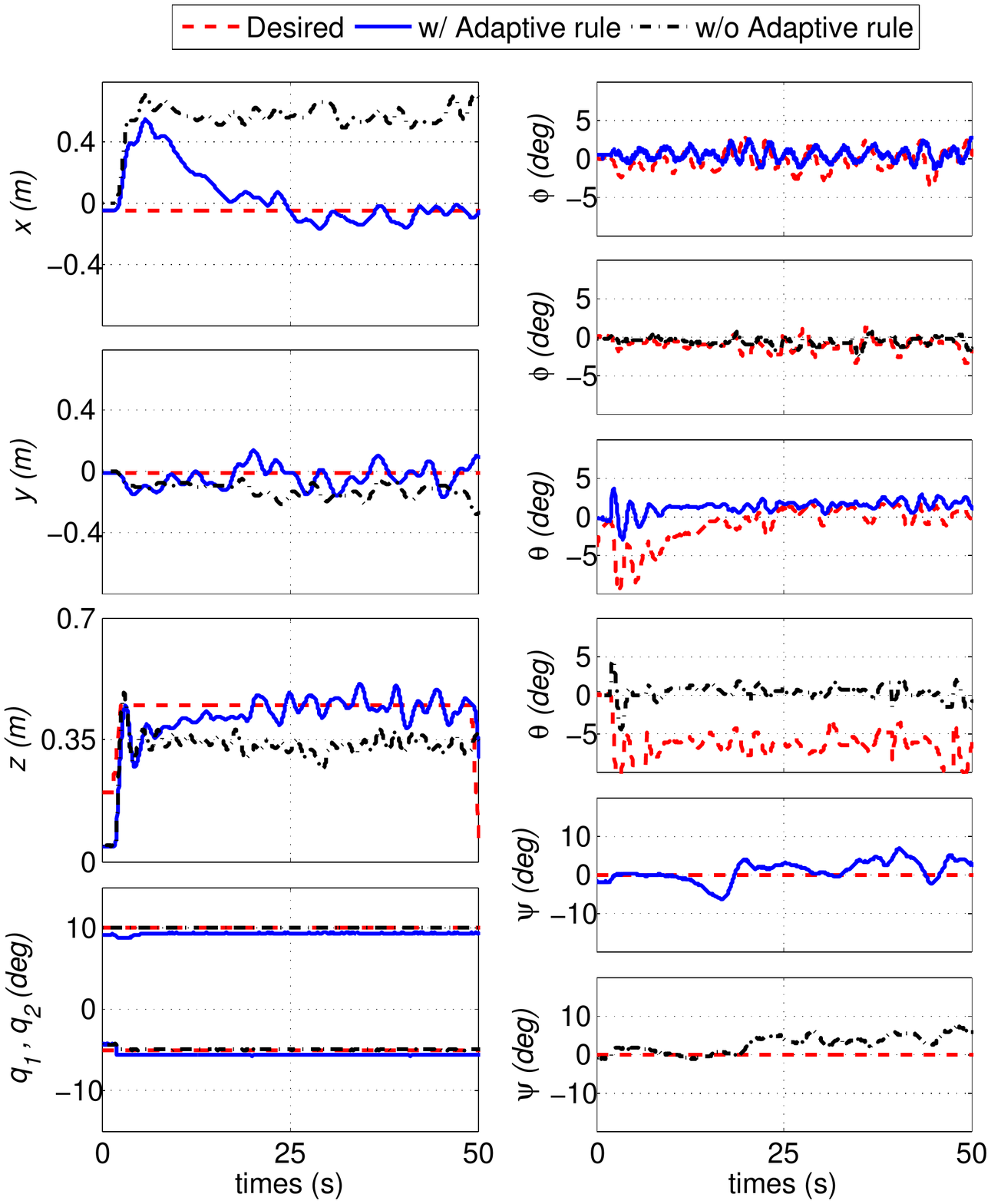}}
    \subfloat[Estimator and parameter estimation results\label{fig:flight2}]
        {\includegraphics[width=0.5\columnwidth]{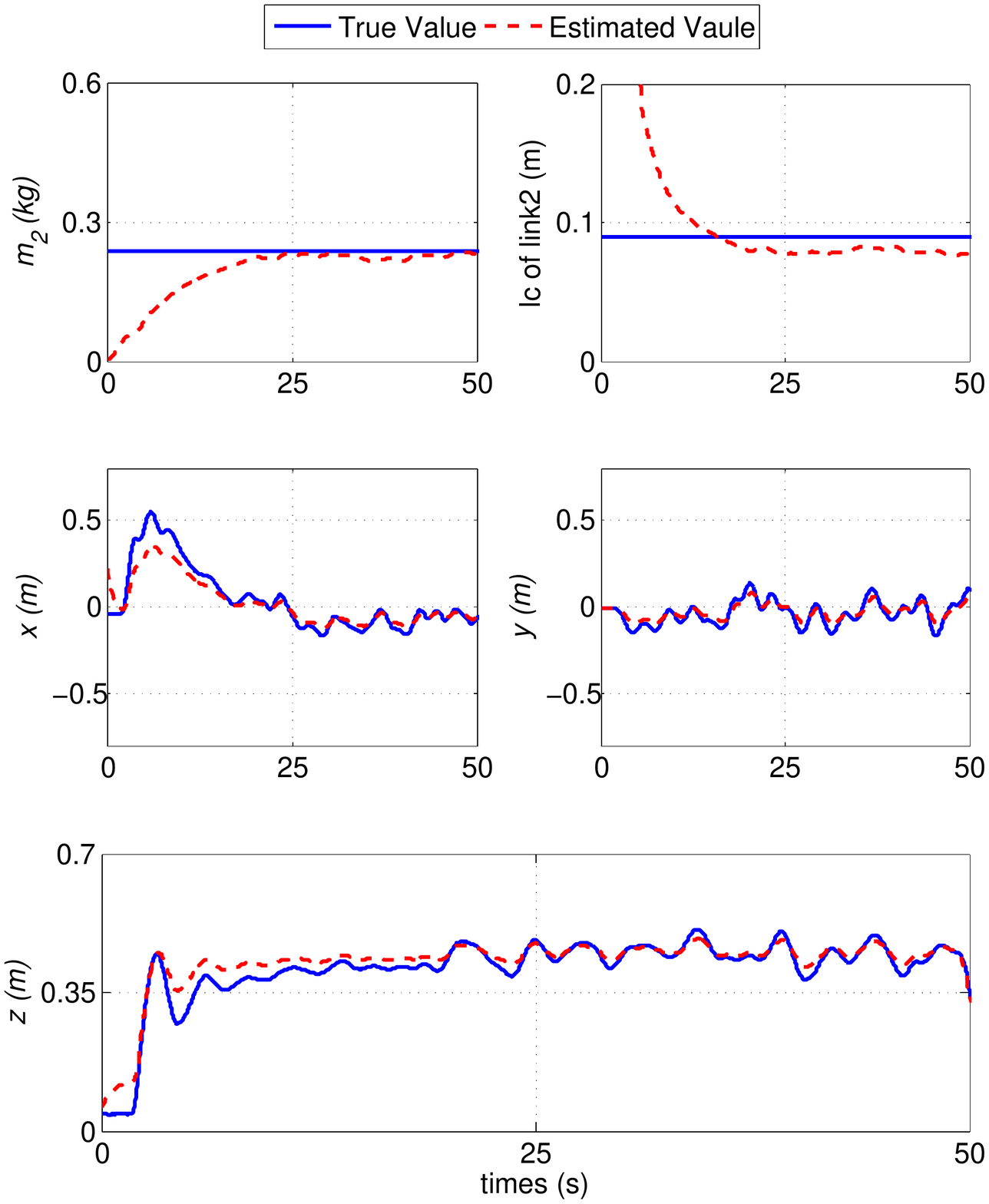}}
    \caption{\textrm{\small Experimental results of carrying on an unknown payload.}}
	\label{fig:adapt_flight}
\end{figure}

The proposed control algorithm using a parameter estimator and controller is validated by an experiment. The flight result is shown in Fig. \ref{fig:adapt_flight}

Fig. \ref{fig:flight1} compares state tracking performance between the proposed controller and conventional non-adaptive controller. The conventional controller means that the controller does not have update rule for the unknown parameter in \eqref{eqn:tau2}. The red dashed lines show the desired trajectory based on each controller, the blue solid lines represent states with the proposed approach and the black dash-dot lines are states with conventional approach. In Fig. \ref{fig:flight1}, the desired joint angles are fixed at $q_1=-5$ degree and $q_2=10$ degree, respectively, (i.e. the arm is almost stretched to the side.) From the entire trajectory histories, although a large torque is applied due to this particular pose, the proposed controller shows satisfactory tracking results. On the other hand, the conventional approach cannot track the desired trajectory because this controller cannot compensate the effect of the unknown payload.

Fig. \ref{fig:flight2} shows estimator performance and parameter estimation results. The blue solid lines are real state of the aerial manipulator and the red dashed lines show the result by parameter estimator. The parameter estimator estimates current trajectory of the aerial manipulator satisfactorily. The parameter estimator shows good convergence to the true mass of $m_2$ at 0.238 kg (the additional payload is 0.120 kg for the extra payload and the gripper) and $l_c$ about 0.09 meter.

A video clip of the experiments is posted on the following URL: \url{http://icsl.snu.ac.kr/hbeom/Estimation_AMS.mp4.} 
\section{Conclusion} \label{sec:con}
This paper presented a parameter estimation and control of an aerial manipulator for handling an additional payload. The unknown parameters of the payload were estimated and an augmented passivity-based controller was designed to control a hexacopter with a 2-DOF robotic arm. In the experimental results, we showed a successful flight using a custom-made aerial manipulator while the unknown parameters were estimated satisfactorily. Our future works include adaptive control of aerial manipulator for handling lumped uncertainty except the unknown payload.


\bibliographystyle{IEEEtran}	
\bibliography{hbeom_case2015}		

\end{document}